\documentclass[10pt,twocolumn,letterpaper]{article}

\usepackage{iccv}
\usepackage{times}
\usepackage{epsfig}
\usepackage{graphicx}
\usepackage{amsmath}
\usepackage{amssymb}
\usepackage{caption}
\usepackage{booktabs}
\usepackage{array,multirow}
\usepackage{soul}
\usepackage{pifont}
\usepackage[numbers,sort]{natbib} 
\usepackage{xspace}
\usepackage{adjustbox}
\usepackage{subcaption}
\usepackage{enumitem}
\usepackage{diagbox}
\usepackage{hhline}
\usepackage[accsupp]{axessibility}
\usepackage{pdfpages}

\DeclareMathSymbol{\mhyphen}{\mathord}{AMSa}{"39}
\newcommand\scalemath[2]{\scalebox{#1}{\mbox{\ensuremath{\displaystyle #2}}}}

\usepackage[pagebackref=true,breaklinks=true,letterpaper=true,colorlinks,bookmarks=false]{hyperref}

\iccvfinalcopy 

\ificcvfinal\fi

\begin{document}

\title{RankMixup: Ranking-Based Mixup Training for Network Calibration}

\author{Jongyoun Noh \quad\quad\quad Hyekang Park \quad\quad\quad Junghyup Lee \quad\quad\quad Bumsub Ham\thanks{Corresponding author.}\vspace*{0.2cm}\\
{School of Electrical and Electronic Engineering, Yonsei University}
\vspace{1mm}\\
\url{https://cvlab.yonsei.ac.kr/projects/RankMixup}}

\maketitle
\ificcvfinal\thispagestyle{empty}\fi

\begin{abstract}
   Network calibration aims to accurately estimate the level of confidences, which is particularly important for employing deep neural networks in real-world systems. Recent approaches leverage mixup to calibrate the network's predictions during training. However, they do not consider the problem that mixtures of labels in mixup may not accurately represent the actual distribution of augmented samples. In this paper, we present RankMixup, a novel mixup-based framework alleviating the problem of the mixture of labels for network calibration. To this end, we propose to use an ordinal ranking relationship between raw and mixup-augmented samples as an alternative supervisory signal to the label mixtures for network calibration. We hypothesize that the network should estimate a higher level of confidence for the raw samples than the augmented ones~(Fig.\ref{fig:teaser}). To implement this idea, we introduce a mixup-based ranking loss~(MRL) that encourages lower confidences for augmented samples compared to raw ones, maintaining the ranking relationship. We also propose to leverage the ranking relationship among multiple mixup-augmented samples to further improve the calibration capability. Augmented samples with larger mixing coefficients are expected to have higher confidences and vice versa~(Fig.\ref{fig:teaser}). That is, the order of confidences should be aligned with that of mixing coefficients. To this end, we introduce a novel loss, M-NDCG, in order to reduce the number of misaligned pairs of the coefficients and confidences. Extensive experimental results on standard benchmarks for network calibration demonstrate the effectiveness of RankMixup.
\end{abstract}
\section{Introduction}

Deep neural networks~(DNNs) have achieved remarkable advances in numerous computer vision tasks, including image classification, object detection, and semantic segmentation. Despite the impressive performance, DNNs trained using a softmax cross-entropy~(CE) loss and one-hot encoded labels often output overconfident and/or underconfident probabilities, which do not reflect the actual likelihood~\cite{guo2017calibration,thulasidasan2019mixup,nguyen2015deep}. This suggests that the networks are not able to estimate the level of confidence, leading to inaccurate and unreliable predictions, which restricts practical applications in real-world systems, particularly for safety-sensitive domains, such as medical diagnosis~\cite{jiang2012calibrating} and autonomous driving~\cite{grigorescu2020survey}.

\begin{figure}[t]
  
   \centering
   \includegraphics[width=1.0\linewidth]{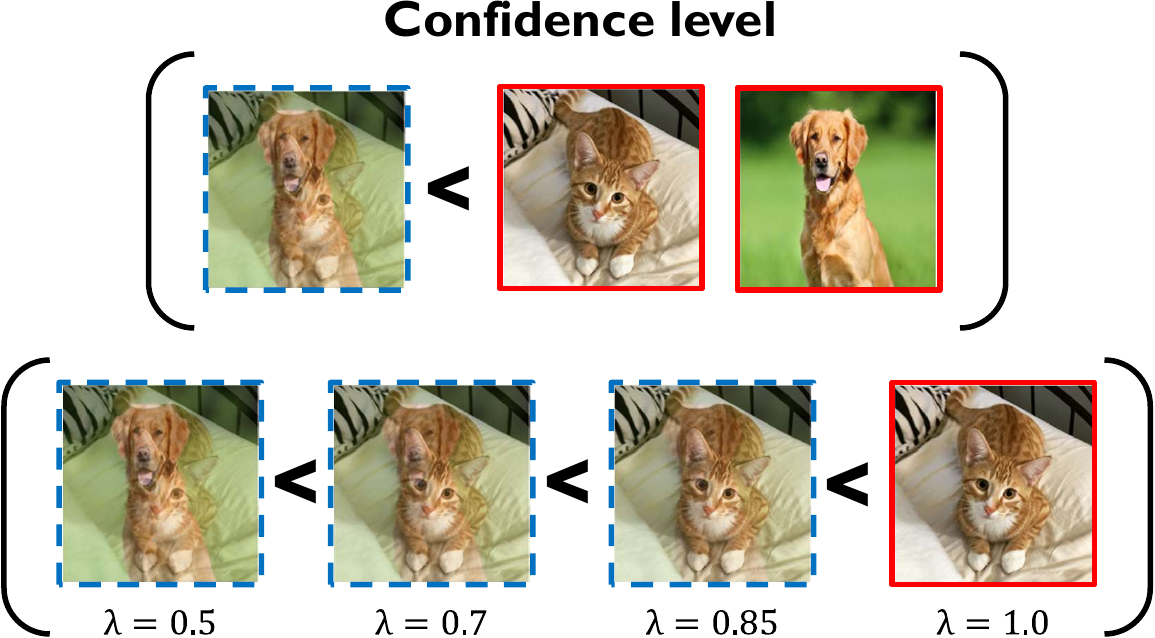}
   \caption*{}

\captionsetup{font={small}}
\vspace{-7mm}
\caption{Motivation of RankMixup. 
We show raw samples of a cat and dog in red boxes, and mixup-augmented samples in blue boxes with dashed lines. The mixup-augmented samples are generated by linearly interpolating the raw samples with a mixing coefficient~$\lambda$. We expect that confidences of the raw samples to be higher than that of the augmented sample~(top). The augmented samples with larger coefficients $\lambda$ would have higher confidences than the samples with smaller coefficients~(bottom). Best viewed in color.}
\label{fig:teaser}
\end{figure}

To improve the reliability of DNNs, many methods propose to calibrate the output distribution after or during training. Post-hoc approaches adjust miscalibrated probabilities directly from pretrained networks~\cite{guo2017calibration,ma2021meta,gupta2020calibration,kull2019beyond,ding2021local}. Although these approaches are straightforward to apply, calibration performance depends heavily on datasets and networks used for training~\cite{liu2022devil}, and they are not robust to distribution shifts between training and test samples~\cite{ovadia2019can}. Training-time approaches~\cite{thulasidasan2019mixup,zhang2022and,ghoshadafocal,cheng2022calibrating,hebbalaguppe2022stitch,liu2022devil,moon2020confidence,muller2019does,mukhoti2020calibrating}, on the other hand, exploit an additional regularization term along with the CE loss to calibrate the output probabilities during training. Early works~\cite{thulasidasan2019mixup,muller2019does,mukhoti2020calibrating} have explored the effects of regularization techniques, including label smoothing~(LS)~\cite{szegedy2016rethinking}, focal loss~(FL)~\cite{lin2017focal}, and mixup~\cite{zhang2018mixup}, on network calibration, and shown that the networks trained with these techniques are better calibrated than the ones trained with the CE loss alone. Recently, the works of~\cite{liu2022devil,ghoshadafocal,mukhoti2020calibrating} highlight the limitations of LS and FL in terms of network calibration, and introduce calibration-aware LS and FL. On the contrary, few studies have explored integrating the mixup technique into network calibration. Current mixup-based methods~\cite{zhang2022and,thulasidasan2019mixup,pinto2022using} simply interpolate input images and training labels with a mixing coefficient to soften output probabilities, while not considering network calibration itself. Since the interpolated labels may not represent accurately the distribution of the label mixture in augmented images~\cite{sun2022lumix,kim2020puzzle,chen2022transmix,uddin2021saliencymix}, training networks with such uncertain labels degrades the calibration performance.

In this work, we present a mixup-based approach to network calibration, dubbed RankMixup, that addresses the problem of directly using the mixture of labels for calibrating network predictions~\cite{zhang2022and,thulasidasan2019mixup,pinto2022using}. To this end, we conjecture that networks should give higher confidences for the raw sample, not augmented by the mixup technique, than the augmented one~(Fig.~\ref{fig:teaser}). To implement this idea, we leverage an ordinal ranking relationship between raw and augmented samples~(Fig.~\ref{fig:teaser}) and use it as a supervisory signal for network calibration, instead of using the label mixtures. In particular, we introduce a mixup-based ranking loss~(MRL) that encourages the confidences of augmented samples to be lower than those of raw samples by a margin, in order to preserve the ranking relationship. To further strengthen the capability of network calibration, we also propose to exploit the ranking relationship for multiple mixup-augmented samples. We expect that higher confidences are favorable for augmented samples with larger mixing coefficients~$\lambda$, and vice versa~(Fig.~\ref{fig:teaser}), such that the orders of confidences and mixing coefficients are aligned to each other. To achieve this, we introduce a novel loss, M-NDCG, based on normalized discounted cumulative gain~(NDCG) in information retrieval~\cite{jarvelin2002cumulated}, which penalizes the number of misaligned pairs of the coefficients and confidences. Extensive experimental results on standard benchmarks demonstrate the effectiveness of our approach to leveraging the ordinal ranking relationships on network calibration. The main contributions of our work can be summarized as follows:
\vspace{-2mm}
\begin{itemize}[leftmargin=*]
   \item[$\bullet$] We introduce a novel mixup-based framework for network calibration, RankMixup, that alleviates the problem of exploiting label mixtures for calibrating network predictions. To the best of our knowledge, we are the first to handle this issue in the context of confidence calibration. 
   \vspace{-2mm}
   \item[$\bullet$] We propose to exploit ordinal ranking relationships between confidences from raw and augmented samples, as well as those among confidences from multiple augmented samples, and use them as supervisory signals, instead of the mixture of labels. To implement this idea, we introduce MRL and M-NDCG that facilitate aligning the confidences and mixing coefficients to preserve the relationships.
   \vspace{-2mm}
   \item[$\bullet$] We demonstrate that our approach provides better results in terms of expected calibration error~(ECE), compared to other mixup-based methods. We also achieve competitive performances on standard benchmarks on network calibration, including CIFAR10/100~\cite{krizhevsky2009learning},  Tiny-ImageNet~\cite{le2015tiny}, and ImageNet~\cite{deng2009imagenet} with various network architectures.
\end{itemize}
\section{Related work}

\paragraph{Confidence calibration.}

The seminal work of~\cite{guo2017calibration} provides empirical evidence that DNNs are prone to making overconfident predictions due to miscalibrated output probabilities. In particular, they are typically trained with one-hot encoded labels, which fails to impose any entropy on the supervisory signal. The entropy of output probabilities for the network is thus encouraged to be zero, leading to the overconfident predictions~\cite{liu2022devil}. To address this problem, several approaches have been proposed to augment the entropy of output probabilities, which can be divided into two categories: post-hoc and training-time methods. First, post-hoc approaches adjust the miscalibrated probabilities directly after training using a temperature scaling technique~\cite{guo2017calibration,ding2021local,tomani2021post} or ranking statistics~\cite{ma2021meta}. The temperature scaling technique, also effective to knowledge distillation~\cite{hinton2015distilling}, raises the entropy of output probabilities using a temperature parameter, optimized on additional validation samples. Although post-hoc approaches have shown the effectiveness on in-distribution datasets, they perform poorly under distribution shifts~\cite{ovadia2019can}, and inevitably lower the confidences of correct predictions~\cite{mukhoti2020calibrating,kumar2018trainable}. Second, training-time approaches implicitly or explicitly impose regularization on outputs probabilities during training, usually in combination with post-hoc methods to further improve the calibration accuracy. The implicit methods~\cite{cheng2022calibrating,thulasidasan2019mixup,muller2019does,liu2022devil,pinto2022using,ghoshadafocal,cheng2022calibrating} smooth the output distribution by increasing the entropy of training labels, rather than adjusting output probabilities directly. Several works~\cite{muller2019does,mukhoti2020calibrating,thulasidasan2019mixup,zhang2022and} have demonstrated empirically and theoretically that regularization methods such  as LS~\cite{szegedy2016rethinking}, FL~\cite{lin2017focal}, and Mixup~\cite{zhang2018mixup}, which have originally designed to learn discriminative features, can also be effective in network calibration. However, recent works of~\cite{liu2022devil,ghoshadafocal,pinto2022using} explore that directly using these methods for network calibration is problematic, and they introduce improved versions to overcome the drawback.  For example, MbLS~\cite{liu2022devil} exploits a margin-based LS to address the problem of LS providing non-informative predictions to obtain a better-calibrated network. The works of~\cite{ghoshadafocal,pinto2022using} suggest that finding optimal parameters of FL and mixup is crucial for network calibration. The explicit methods~\cite{pereyra2017regularizing,moon2020confidence,hebbalaguppe2022stitch,kumar2018trainable} adjust confidences of predictions directly by maximizing the entropy of output probabilities~\cite{pereyra2017regularizing}, using an ordinal ranking relationship~\cite{moon2020confidence} or minimizing calibration errors~\cite{kumar2018trainable,hebbalaguppe2022stitch}. For example, the ordinal ranking between output probabilities of networks has been proven to be effective to estimate the levels of confidence, providing better calibrated models~\cite{moon2020confidence}. It assumes that the predictions with higher confidences are more likely to be accurate than the ones with lower confidences. 
We explore the effects of mixup~\cite{zhang2018mixup} in the context of network calibration, similar to other mixup-based methods~\cite{thulasidasan2019mixup,zhang2022and, pinto2022using}. They use interpolated labels to train networks for calibrating output probabilities, but the labels do not provide the distribution of the label mixture for the augmented sample accurately~\cite{sun2022lumix,kim2020puzzle,chen2022transmix,uddin2021saliencymix}. We exploit confidences of augmented samples directly as supervisory signals for network calibration using ordinal ranking relationships between the samples, boosting the calibration performance significantly. The work of~\cite{moon2020confidence} is closely related to ours in that it uses an ordinal ranking relationship for raw samples. In contrast to this, our approach exploits more diverse and complex ordinal relationships for confidences. It considers not only the relationship between raw and augmented samples, but also among the augmented samples themselves.
\vspace{-4mm}
\paragraph{Mixup.}
Data augmentation has become a standard technique for training DNNs that regularizes the networks to prevent overfitting and improve generalization. Mixup~\cite{zhang2018mixup} linearly interpolates between pairs of raw data samples. Specifically, it takes a weighted average of two images to obtain a new one. The training label for the augmented sample is also interpolated with the same weight. Many works extend this idea to improve the generalization ability. Manifold-Mixup~\cite{verma2019manifold} performs the mixup process in the feature space, while CutMix~\cite{yun2019cutmix} introduces a regional mixing scheme to preserve fine-grained information of original images. It has recently been explored that using label mixtures in mixup and its variants could provide misleading supervisory signals for training networks, which can be alleviated by saliency-guided mixup schemes~\cite{chen2022transmix,uddin2021saliencymix,qin2020resizemix,kim2020puzzle} or label perturbations~\cite{sun2022lumix}. In addition to data augmentation, various applications leverage mixup for,~\eg,~self-supervised learning~\cite{shen2022mix,ren2022simple} and network calibration~\cite{thulasidasan2019mixup,pinto2022using}. We mainly investigate the effect of the vanilla mixup~\cite{zhang2018mixup} on network calibration. However, as will be shown in our experiments, the variants of mixup, including CutMix~\cite{yun2019cutmix} and manifold-Mixup~\cite{verma2019manifold}, can also be readily adapted to our approach.

\section{Approach}
We present in this section preliminaries for network calibration and mixup~\cite{zhang2018mixup} in Sec.~\ref{sec:pre}, and provide a detailed description of RankMiup and training losses in Secs.~\ref{sec:rankmixup} and~\ref{sec:loss}, respectively.

\subsection{Preliminaries}\label{sec:pre}
\paragraph{Network calibration.}
Let us suppose a dataset~$D = \{ ( \mathbf{x}_{i}, \mathbf{y}_{i} ) \}_{i=1}^N$ of size $N$, where $\mathbf{x}_i \in \mathcal{X}$ and $\mathbf{y}_i\in \mathcal{Y}$ are an input sample and a ground-truth label with $K$ classes, respectively. DNNs can be considered as a function, parameterized by $\theta$, that yields a logit $\mathbf{z}_{i}$ of size $K\times 1$ from the input sample: $\mathbf{f}_{\theta}:\mathcal{X} \to \mathbb{R}^{K}$. For multiclass classification, the networks compute class probabilities using the softmax function as follows: 
\begin{equation}\label{eq:p}
   {p}_{i,k} = \mathrm{softmax}(\mathbf{z}_{i}) = \frac{\exp{z}_{i,k}}{\sum_{j}^{K} \exp{z}_{i,j}}, 
\end{equation}
where we denote by ${p}_{i,k}$ a probability of the $k$-th class for the $i$-th input. Based on the probabilities, the networks predict a class for the input sample~$\mathbf{x}_{i}$ as
\begin{equation}
	\hat{{y}}_{i} = \mathrm{arg}\max_{k} {p}_{i,k}.
\end{equation}
The corresponding confidence can then be obtained as follows:
\begin{equation}
 \hat{{p}}_{i} 	= \max_{k} {p}_{i,k}.
\end{equation}
During training, the networks are optimized using a cross-entropy~(CE) loss as follows:
\begin{equation}\label{eq:ce}
   \mathcal{L}_{CE}(\mathbf{x}_{i},\mathbf{y}_{i}) = -\sum_{k=1}^{K} {y}_{i,k}\log{{p}_{i,k}},
\end{equation} 
where ${y}_{i,k}$ is the $k$-th element of the one-hot encoded label~$\mathbf{y}_{i}$. Recent works~\cite{guo2017calibration,muller2019does,thulasidasan2019mixup,mukhoti2020calibrating} have shown that DNNs trained with the CE loss tend to be overconfident, providing higher confidences than accuracies. To address this problem, confidence calibration methods encourage that the confidence for each sample and the accuracy are the same as follows: 
\begin{equation}
\hat{{p}}_{i} = P(\hat{{y}}_{i}={y}_{i} | \hat{{p}}_{i}),	
\end{equation}
where $y_i$ is an index of the true class for the ground-truth label~$\mathbf{y}_i$.

\paragraph{Mixup.}
Mixup~\cite{zhang2018mixup} is a data-augmentation technique based on the vicinal risk minimization principle~\cite{chapelle2000vicinal}. It introduces the mixup vicinal distribution, which measures the likelihood of finding a pair of an augmented input and a target label,~$(\tilde{\mathbf{x}}_{i}, \tilde{\mathbf{y}}_{i})$, from the vicinity of the original pair,~$(\mathbf{x}_{i}, \mathbf{y}_{i})$. Mixup samples the vicinal distribution to construct a new dataset, $D_{v} = \{ ( \tilde{\mathbf{x}}_{i}, \tilde{\mathbf{y}}_{i} ) \}_{i=1}^N$. Here, we denote by $D_{v}$ the vicinal set consisting of pairs of augmented training samples and corresponding target labels. Specifically, it augments training pairs by linearly interpolating the original ones as follows:

\begin{equation}\label{eq:mixup}
   \begin{aligned}
   \tilde{\mathbf{x}} = \lambda \mathbf{x}_{i} + (1 - \lambda)\mathbf{x}_{j}, \\
   \tilde{\mathbf{y}} = \lambda \mathbf{y}_{i} + (1 - \lambda)\mathbf{y}_{j},
   \end{aligned}   
\end{equation}
where two pairs of input samples and target labels, $(\mathbf{x}_{i}$, $\mathbf{y}_{i})$ and $(\mathbf{x}_{j}$, $\mathbf{y}_{j})$, are randomly drawn from the training set~$D$, and $\lambda\in[0, 1] \sim \text{Beta}(\alpha, \alpha)$ is a mixing coefficient, sampled from a Beta distribution. Mixup assumes that target labels for the vicinal set should also be interpolated with the same mixing coefficient as input samples during the mixing process. The hyperparameter~$\alpha$ controls the strength of interpolation between the pairs of input samples and target labels. According to~\cite{zhang2018mixup}, the value of $\alpha$ within the range of $[0.1, 0.4]$ provides better classification performance, while higher values cause significant underfitting. We will also examine the effect of $\alpha$ in terms of confidence calibration within our framework.

\subsection{RankMixup}\label{sec:rankmixup}
Mixup provides smooth decision boundaries that gradually shift from one class to another, which leads to a regularized estimate of uncertainty~(or confidence)~\cite{zhang2018mixup}. There have been lots of studies to discover underlying principles behind the mixup method~\cite{zhang2022and,zhang2021how,thulasidasan2019mixup,zhang2022and,hendrycksaugmix}. For example, it has proven to give better output probabilities in terms of model calibration, alleviating overconfident predictions~\cite{thulasidasan2019mixup,pinto2022using,zhang2022and}. The interpolated labels of augmented samples in the vicinal set do not always model the actual distribution of label mixtures in augmented samples~\cite{sun2022lumix,kim2020puzzle,chen2022transmix,uddin2021saliencymix}, which however has not been considered for designing a network calibration method. To alleviate this problem, we leverage ordinal ranking relationships between confidences from raw and augmented samples, as well as those among confidences from multiple augmented samples. We then propose to use them as supervisory signals for model calibration, rather than using the mixture of labels directly. In the following, we describe our framework with a single mixup-augmented sample, and present a generalized version that can handle multiple augmented samples.  

\paragraph{Single mixup-augmented sample.}
The sample augmented by mixup~$\tilde{\bf{x}}$ in Eq.~\eqref{eq:mixup} can be interpreted as a combination of  the raw sample~${\bf{x}}_{i}$ and structured noise from another one~${\bf{x}}_{j}$, where the degree of noise is determined by the mixing parameter~$\lambda$. This suggests that DNNs predict the class of the augmented sample with uncertainty imposed by the additional one, making it more difficult to recognize the augmented sample than the raw one. To this end, mixup imposes uncertainty on the label mixtures by adjusting confidences of target labels for the augmented samples by~$\lambda$~(or~$1-\lambda$). Albeit simple, directly using the interpolated labels may provide misleading supervisory signals to train DNNs. To avoid this, we leverage an ordinal ranking relationship between confidences of raw and augmented samples. Specifically, the ordinal ranking holds for the relationship that confidence levels of easy-to-classify samples should be higher than the ones for hard samples~\cite{moon2020confidence}. Accordingly, we conjecture that the confidence levels of raw samples should be greater than those of the augmented samples in the mixing process~(Fig.\ref{fig:teaser}). Namely, every pair of samples~$(\mathbf{x}_{i}, \tilde{\mathbf{x}}_{i})$ should satisfy the following relationship:  
\begin{equation}\label{eq:sinrank}
   \max_{k} {p}_{i,k} \geq \max_{k} \tilde{{p}}_{i,k},
\end{equation}
where $\tilde{{p}}_{i,k}$ is a predicted probability of the class~$k$ for the augmented sample $\tilde{\mathbf{x}}_{i}$. We devise MRL that encourages the network to learn this relationship without the mixture of labels, which will be described in Sec.~\ref{sec:loss}.

\paragraph{Multiple mixup-augmented samples.}
We extend our framework to exploit multiple mixup-augmented samples. As the mixing ratio~$\lambda$ decreases, augmented samples become more difficult to recognize, which suggests that we can exploit more complex ranking relationships among the multiple augmented samples. Analogous to the single case, we expect that the samples augmented with larger mixing coefficients~$\lambda$ show higher confidences compared to those with smaller coefficients~(Fig.~\ref{fig:teaser}). For every pair of augmented samples $( \tilde{\mathbf{x}}_{i}, \tilde{\mathbf{x}}_{j} )$ from the vicinal set~$D_{v}$, the confidences of these samples should satisfy the following relationship:
\begin{equation}\label{eq:mulrank}
   \lambda_{i} \geq \lambda_{j} 
   \Leftrightarrow \max_{k} \tilde{{p}}_{i,k} \geq  \max_{k} \tilde{{p}}_{j,k},
\end{equation}
where $(\lambda_{i}, \lambda_{j})$ is a pair of mixing ratios for the corresponding pair of augmented samples~$(\tilde{\mathbf{x}}_{i}, \tilde{\mathbf{x}}_{j})$. Note that we describe the relationship with two samples for simplicity, but our framework can adopt the relationship with any number of augmented samples.

We then integrate both the relationships in Eqs.~\eqref{eq:sinrank} and~\eqref{eq:mulrank} to obtain a general version of RankMixup. With the mixing coefficient of~$1.0$ for the raw sample, the following relationship should hold for every triplet of samples~$(\mathbf{x}_{i}, \tilde{\mathbf{x}}_{i}, \tilde{\mathbf{x}}_{j})$\footnote{Note that Eq.~\eqref{eq:rankmixup} holds only when $\lambda \geq 0.5$.}: 
\begin{equation}\label{eq:rankmixup}
1.0 \geq \lambda_{i} \geq \lambda_{j} 
\Leftrightarrow \max_{k} {p}_{i,k} \geq \max_{k} \tilde{{p}}_{i,k} \geq  \max_{k} \tilde{{p}}_{j,k}.
\end{equation}
This allows the network to estimate the levels of confidence better by considering diverse ordinal ranking relationships among the confidences of the raw and multiple mixup-augmented samples. To implement this idea, we align the orders of confidences and mixing ratios during training using M-NDCG~(Sec.~\ref{sec:loss}).

\subsection{Training loss}\label{sec:loss}
To train our model, we use a CE loss for classification, together with either MRL or M-NDCG for network calibration, balanced by the parameter of~$w$. As supervisory signals, we exploit one-hot encoded labels for classification, and ordinal ranking relationships for input and augmented samples for network calibration.

\paragraph{MRL.}
We introduce MRL facilitating the ordinal relationship in Eq.~\eqref{eq:sinrank}. It encourages confidences of augmented samples to be lower than those of raw samples by a margin, defined as follows:
\begin{equation}\label{eq:mrl}
   \mathcal{L}_\text{MRL} = \max(0, \max_{k} \tilde{{p}}_{i,k} - \max_{k} {p}_{i,k} + {m}),
\end{equation}
where the margin~$m$ determines acceptable discrepancies between the confidences of raw and augmented samples. The intuition behind using a margin is twofold: (1) It enables maintaining the ranking relationship between confidences, rather than penalizing the exact value of the discrepancies itself. (2) Controlling the confidence discrepancies with a margin allows to consider the uncertainty of inaccurate label mixtures. The vanilla mixup method continues to penalize the confidences of augmented samples, until they reach overconfident values, determined by the label mixtures. In contrast to this, MRL penalizes the confidences, only when the absolute discrepancies between them are below the margin.

\paragraph{M-NDCG.}
MRL is not sufficient to model the relationship in Eq.~\eqref{eq:rankmixup}, where the comparison involves more than one augmented sample. M-NDCG addresses this problem that enables multiple confidence values to hold the ordinal ranking relationships. M-NDCG is based on NDCG used for evaluating ranking quality in information retrieval~\cite{jarvelin2002cumulated, valizadegan2009learning}. It measures how many of retrieved results are aligned with the ground-truth ranking order, which can be calculated by normalizing a discounted cumulative gain~(DCG) with an ideal DCG~(IDCG):
\begin{equation}\label{eq:ndcg}
   NDCG = \frac{DCG}{IDCG}.
\end{equation}
DCG accumulates a relevance score of each retrieval result, weighted by a corresponding rank, and IDCG is computed similarly but with the ground-truth score, as follows: 
\begin{equation}\label{eq:dcg}
   \begin{aligned}
   DCG &=\sum_{q=1}^{Q}\frac{r_{q}}{\log_{2}(q+1)}, \\
   IDCG &=\sum_{q=1}^{Q}\frac{r_{q}^\text{gt}}{\log_{2}(q+1)},
   \end{aligned}
\end{equation}
where $r_q$ and $r_{q}^\text{gt}$ are the relevance and ground-truth scores, respectively, at position~$q$ in the result list of size $Q$.

We consider confidences of predictions and mixing coefficients as the relevance and ground-truth scores, respectively, enabling multiple confidences to satisfy the relationship in Eq.~\eqref{eq:rankmixup}. Given $Q$ pairs of confidences and mixing coefficients, we compute $DCG_{\text{M}}$ and $IDCG_{\text{M}}$, as follows:  
\begin{equation}\label{eq:mdcg}
   \begin{aligned}
   DCG_{\text{M}} &=\frac{\max_{k} {p}_{t,k}}{\log_{2}(t+1)} + \sum_{q=1,q\neq t}^{Q}\frac{\max_{k} \tilde{{p}}_{q,k}}{\log_{2}(q+1)}, \\
   IDCG_{\text{M}}&= 1 + \sum_{q=2}^{Q}\frac{\lambda_{q}^\text{gt}}{\log_{2}(q+1)}, 
   \end{aligned}
\end{equation}
where $t$ is the position of the confidence value for the raw sample, which is set to 1 in order to ensure that its confidence is always ranked first in a descending order, following the ranking relationship in Eq.~\eqref{eq:rankmixup}. $Q$~is a hyperparameter that determines the number of augmented samples, \ie,~$Q-1$ augmented samples. $\lambda_{q}^\text{gt}$ is the mixing coefficient sorted based on its magnitude, with a value of 1.0 for the raw sample appearing at position of the top rank,~\ie,~$\lambda_{1}^\text{gt}=1$. We then define M-NDCG loss as follows:
\begin{equation}\label{eq:mndcg}
   \mathcal{L}_{\text{M-NDCG}} = 1 - \frac{DCG_\text{M}}{IDCG_\text{M}},
\end{equation}    
which is minimized when the ranking orders of the confidences and mixing coefficients are aligned to each other after sorting in a descending ordeer. The network is thus trained to encourage the confidences to hold the ordinal ranking relationship of RankMixup in Eq.~\eqref{eq:rankmixup}. M-NDCG also discounts lower-ranked confidences more, preventing the uncertain predictions from being overconfident.

\begin{table*}[]
   \vspace{-2mm}

   \centering
   \captionsetup{font={small}}
   \caption{Quantitative comparison with the state of the art in terms of ECE~(\%) and AECE~(\%) with 15 bins on the validation splits of CIFAR10/100~\cite{krizhevsky2009learning} and Tiny-ImageNet~\cite{le2015tiny}. Numbers in bold indicate the best performance and underscored ones indicate the second best. Numbers in parentheses represent the results obtained using a TS post-hoc technique~\cite{guo2017calibration}.}
   \vspace{-2mm}
   \label{tab:result}
   \setlength{\tabcolsep}{1.2pt}
   \renewcommand{\arraystretch}{1.1}
   \resizebox{\textwidth}{!}{%
   {\scriptsize
   \begin{tabular}{l|ccc|ccc|ccc|ccc|ccc|ccc|}
   \cline{2-19}
    & \multicolumn{6}{c|}{CIFAR10~\cite{krizhevsky2009learning}} & \multicolumn{6}{c|}{CIFAR100~\cite{krizhevsky2009learning}} & \multicolumn{6}{c|}{Tiny-ImageNet~\cite{le2015tiny}}  \\ \hline
   \multicolumn{1}{|c|}{\multirow{2}{*}{Methods}} &
     \multicolumn{3}{c|}{ResNet-50~\cite{he2016deep}} &
     \multicolumn{3}{c|}{ResNet-101~\cite{he2016deep}} &
     \multicolumn{3}{c|}{ResNet-50~\cite{he2016deep}} &
     \multicolumn{3}{c|}{ResNet-101~\cite{he2016deep}} &
     \multicolumn{3}{c|}{ResNet-50~\cite{he2016deep}} &
     \multicolumn{3}{c|}{ResNet-101~\cite{he2016deep}}  \\ \cline{2-19} 
   \multicolumn{1}{|c|}{} &
     \multicolumn{1}{c|}{Acc$\uparrow$} &
     \multicolumn{1}{c|}{ECE$\downarrow$} &
     \multicolumn{1}{c|}{AECE$\downarrow$} &
     \multicolumn{1}{c|}{Acc$\uparrow$} &
     \multicolumn{1}{c|}{ECE$\downarrow$} &
     \multicolumn{1}{c|}{AECE$\downarrow$} &
     \multicolumn{1}{c|}{Acc$\uparrow$} &
     \multicolumn{1}{c|}{ECE$\downarrow$} &
     \multicolumn{1}{c|}{AECE$\downarrow$} &
     \multicolumn{1}{c|}{Acc$\uparrow$} &
     \multicolumn{1}{c|}{ECE$\downarrow$} &
     \multicolumn{1}{c|}{AECE$\downarrow$} &
     \multicolumn{1}{c|}{Acc$\uparrow$} &
     \multicolumn{1}{c|}{ECE$\downarrow$} &
     \multicolumn{1}{c|}{AECE$\downarrow$} &
     \multicolumn{1}{c|}{Acc$\uparrow$} &
     \multicolumn{1}{c|}{ECE$\downarrow$} &
     \multicolumn{1}{c|}{AECE$\downarrow$}\\ \hline 
     \multicolumn{1}{|l|}{CE}                              &  \bf{95.38}             & 3.75(0.97)             & 2.98(1.01)             & 94.46             & 3.61(0.92) & 3.55(\underline{0.85}) &  77.81  & 13.59(2.93) & 13.54(2.86) & 77.48 & 12.94(2.63) & 12.94(2.66) & 64.34              & 3.18(3.18)                  & 2.87(2.87)      & 66.04              & 3.50(3.50)             & 3.52(3.52)\\
     \multicolumn{1}{|l|}{MMCE~\cite{kumar2018trainable}  }                            &  \underline{95.18}              & 3.88(0.97)                  & 3.88(1.12)      & 94.99              & 3.88(1.15)             & 3.88(12.9) & 77.56 & 12.72(2.83) & 12.71(2.86)  & \underline{77.82} & 13.43(3.06) & 13.42(2.80) & 64.80             & 2.03(2.03) & 1.97(1.97) & \bf{66.44}             & 3.40(3.40) & 3.38(3.38) \\
     \multicolumn{1}{|l|}{ECP~\cite{pereyra2017regularizing}   }                       & 94.75              & 4.01(1.06)                  & 3.99(1.53)     & 93.97              & 4.41(1.72)             & 4.40(1.70)  & 76.20 & 12.29(2.08) & 12.28(2.22) & 76.81 & 13.43(2.92) & 13.42(3.04) & 64.88             & 1.94(1.94) & 1.95(1.95) & 66.20             & 2.72(2.72) & 2.70(2.70) \\
     \multicolumn{1}{|l|}{LS~\cite{muller2019does} }                            & 94.87              & 3.27(1.58)                  & 3.67(3.02)      & 94.18              & 3.35(1.51)             & 3.85(3.10)  &  76.45 & 6.73(4.23) & 6.54(4.26) & 76.91 & 7.99(4.38) & 7.87(4.55) & 65.46        & 3.21(2.51) & 3.23(2.51) & 65.52             & 3.11(2.51) & 2.92(2.72)   \\
     \multicolumn{1}{|l|}{FL~\cite{lin2017focal} }                                     &   94.82              & 3.42(1.07)                  & 3.41(0.87)      & 93.59              & 3.27(1.12)             & 3.23(1.37)  & 76.41 & \textbf{2.83}(\underline{1.66}) & \underline{2.88}(\underline{1.73}) & 76.12 & \textbf{3.10}(2.58) & \textbf{3.22}(2.51) & 63.08            & 2.03(2.03) & 1.94(1.94) & 64.02             & 2.18(2.18) & 2.09(2.09)   \\
     \multicolumn{1}{|l|}{Mixup~\cite{thulasidasan2019mixup}}                                      &  94.76             & 2.86(1.37)             & 2.81(2.00)             & 95.50           & 6.87(1.18) & 6.79(2.33)  &  \bf{78.47} & 8.68(2.14) & 8.68(2.19) & \bf{78.74} & 8.92(3.69) & 8.91(3.65) & \bf{65.81}              & 1.92(1.92)                  & 1.96(1.96)      & \underline{66.41}              & 2.41(1.97)             & 2.43(1.95) \\
     \multicolumn{1}{|l|}{FLSD~\cite{mukhoti2020calibrating}}                          &  94.77              & 3.86(\underline{0.83})                  & 3.74(0.96)      & 93.26              & 3.92(0.93)             & 3.67(0.94) &  76.20 & \underline{2.86}(2.86) & \textbf{2.86}(2.86) & 76.61 & \underline{3.29}(\underline{2.04}) & \underline{3.25}(\underline{1.78}) & 63.56             & 1.93(1.93) & 1.98(1.98) & 64.02             & 1.85(1.85) & \underline{1.81}(1.81)  \\
     \multicolumn{1}{|l|}{CRL~\cite{moon2020confidence} }                              &  95.08              & 3.14(0.96)                  & 3.11(1.25)      & 95.04              & 3.74(1.12)             & 3.73(2.03)  &  \underline{77.85} & 6.30(3.43) & 6.26(3.56) & 77.60 & 7.29(3.32) & 7.14(3.31) & 64.88             & 1.65(2.35) & \underline{1.52}(2.34) & 65.87             & 3.57(\underline{1.60}) & 3.56(\textbf{1.52})   \\
     \multicolumn{1}{|l|}{CPC~\cite{cheng2022calibrating}}                             &  95.04              & 5.05(1.89)                  & 5.04(2.60)      & \bf{95.36}  & 4.78(1.52)             & 4.77(2.37) &  77.23 & 13.29(3.74) & 13.28(3.82) & 77.50 & 13.32(2.96) & 13.28(3.23) & \underline{65.70} & 3.41(3.41) & 3.42(3.42) & \bf{66.44}        & 3.93(3.93) & 3.74(3.74)  \\
     \multicolumn{1}{|l|}{MbLS~\cite{liu2022devil}    }                                &  95.25  & \underline{1.16}(1.16)      & 3.18(3.18)      & \underline{95.13}              & \textbf{1.38}(1.38) & 3.25(3.25)  & 77.92 & 4.01(4.01) & 4.14(4.14) & 77.45 & 5.49(5.49) & 6.52(6.52) &  64.74             & \underline{1.64}(\underline{1.64}) & 1.73(1.73) & 65.81             & \underline{1.62}(1.62) & \textbf{1.68}(1.68)  \\ 
     \multicolumn{1}{|l|}{RegMixup~\cite{pinto2022using}  }                             &  94.68              & 2.76(0.98)             & 2.67(0.92)             & 95.03             & 4.75(\underline{0.92}) & 4.74(0.94) & 76.76 & 5.50(1.98) & 5.48(1.98) & 76.93 & 4.20(1.36) & 4.15(1.92) & 63.58             & 3.04(1.89)                  & 3.04(1.81)      & 63.26              & 3.35(1.86)             & 3.32(1.68)  \\ \hline
     \multicolumn{1}{|l|}{RankMixup~(MRL)   }                                                     &  94.59              & \textbf{1.01}(0.84)        & \textbf{1.03}(\underline{0.80}) & 94.18             & \underline{2.86}(1.08) & \textbf{2.87}(1.58) & 76.65             & 3.52(2.40) & 3.39(2.05) & 77.26&4.41(1.68) & 4.24(1.76) & 64.54         & 1.70(1.70)             & 1.69(\underline{1.69}) & 64.80             & 2.26(1.78)        & 2.28(\underline{1.60})  \\ 
     \multicolumn{1}{|l|}{RankMixup~(M-NDCG) }                                                       &  94.88              & 2.59(\bf{0.57})        & \underline{2.58}(\bf{0.52}) & 94.25            & 3.24(\bf{0.65}) & \underline{3.21}(\bf{0.56}) &  77.11             & 3.46(\bf{1.49}) & 3.45(\bf{1.42}) & 76.46& 3.49(\bf{1.10}) & 3.49(\bf{1.40}) & 64.97         & \bf{1.49}(1.49)             & \bf{1.44}(\bf{1.44}) & 64.89              & \bf{1.57}(1.57)        & 1.94(1.94) \\ \hline
   \end{tabular}%
   }
   }
   \vspace{-2mm}

   \end{table*}

\section{Experiments}
\subsection{Implementation details}
\paragraph{Datasets.}
We mainly perform experiments on CIFAR10/100~\cite{krizhevsky2009learning}, Tiny-ImageNet~\cite{le2015tiny} for network calibration. The CIFAR10/100 datasets consist of 50,000 training and 10,000 test images of size 32$\times$32, and provide 10/100 object categories, respectively. Tiny-ImageNet includes 120,000 images of size 64$\times$64 for 200 object categories. Each category in Tiny-ImageNet contains 600 images, which are divided into 500, 50 and 50 images for training, validation and test, respectively. Tiny-ImageNet is more challenging, compared to CIFAR10/100, as it contains lots of object classes with a smaller number of samples per category. We further validate the generalization ability of our method on ImageNet~\cite{deng2009imagenet}, which comprises around 1.2M training and 50K validation images across 1K classes. For out-of-distribution~(OOD) detection, we train a model using CIFAR10, and use CIFAR100, Tiny-ImageNet, and SVHN~\cite{netzer2011reading} as OOD datasets, following the experimental protocol in~\cite{liu2020simple,pinto2022using}. Similarly, CIFAR100 is used to train a model, and evaluate it on CIFAR10, Tiny-ImageNet, and SVHN. 

\vspace{-1mm}
\paragraph{Evaluation protocols.}
Following the standard protocol~\cite{thulasidasan2019mixup,mukhoti2020calibrating,muller2019does,liu2022devil,cheng2022calibrating,hebbalaguppe2022stitch}, we use expected calibration error~(ECE) and adaptive ECE~(AECE) as evaluation metrics for network calibration. ECE measures the expected absolute difference between confidences and corresponding accuracies:
\begin{equation}
    \mathbb{E}\left[\left|P(\hat{{y}}_{i}={y}_{i} | \hat{{p}}_{i})-\hat{p}_{i}\right|\right].	
\end{equation}
To compute ECE, we divide a range $(0, 1]$ into equidistance bins, and compute an average of the absolute differences, weighted by the size of each bin. We set the number of bins for both ECE and AECE as 15, following the protocol. We also report top-1 classification accuracies to compare discriminative capabilities. We provide results obtained without/with applying a post-hoc technique. In particular, we use the TS method~\cite{guo2017calibration}, where the temperature is optimized on the validation sets. For OOD detection, we compare the area under the receiver operating characteristics~(AUROC), following the standard evaluation protocol~\cite{liu2020simple,pinto2022using}.

\vspace{-2mm}
 
\paragraph{Methods for comparison.}
We compare our method with various training-time approaches. In addition to the baseline model using the CE loss, we include implicit methods~(LS~\cite{muller2019does}, FL/FLSD~\cite{mukhoti2020calibrating}, Mixup~\cite{thulasidasan2019mixup}, mbLS~\cite{liu2022devil}, and RegMixup~\cite{pinto2022using}) and explicit methods~(ECP~\cite{pereyra2017regularizing}, MMCE~\cite{kumar2018trainable}, and CPC~\cite{cheng2022calibrating}). We have reproduced results using official codes with hyperparameter settings provided by the authors, if they are not available. Detailed descriptions for the hyperparameters can be found in the supplement.
 
\vspace{-2mm}

\paragraph{Training details.}
 Following~\cite{liu2022devil}, we train our network for 200 epochs for CIFAR10/100~\cite{krizhevsky2009learning}, with an initial learning rate of 1e-1 and a batch size of 128. We also follow the training setting in~\cite{liu2022devil} for Tiny-ImageNet~\cite{le2015tiny} except for a batch size of 32 due to memory limits. All networks are trained using the SGD optimizer with a momentum of 0.9. The learning rate decreases by a tenth at every fixed intervals. We use ResNet50~\cite{he2016deep}, and ResNet101~\cite{he2016deep} for both CIFAR10/100~\cite{krizhevsky2009learning} and Tiny-ImageNet~\cite{le2015tiny}, and ResNet50 for ImageNet~\cite{deng2009imagenet}. For more comprehensive experiments, we also exploit WideResNet~\cite{zagoruyko2016wide} for CIFAR10 and Tiny-ImageNet. We fix the balancing parameter of training loss $w$ to 0.1. We set margin~$m$ to 2.0 and 1.0 for CIFAR10/100 and Tiny-ImageNet, respectively, using each validation dataset. Other hyperparameters, such as $\alpha$ and $Q$, are chosen similarly. For OOD detection, we train the network using M-NDCG and in-distribution~(ID) datasets with the same training settings used for network calibration.

   \begin{table}[]
      \vspace{-2mm}
      \centering
      \captionsetup{font={small}}
      \caption{Quantitative comparison with the state of the art in terms of ECE~(\%) and AECE~(\%) with 15 bins on the validation split of ImageNet~\cite{deng2009imagenet}. Numbers in bold indicate the best performance and underscored ones indicate the second best.}
      \vspace{-2mm}

      \label{tab:imagenet}
      \renewcommand{\arraystretch}{1.0}
{\scriptsize
  \begin{tabular}{l|ccc|}
   \cline{2-4}
   & \multicolumn{3}{c|}{ImageNet~\cite{deng2009imagenet}}  \\ \hline
   \multicolumn{1}{|c|}{\multirow{2}{*}{Methods}} &
     \multicolumn{3}{c|}{ResNet-50~\cite{he2016deep}}   \\ \cline{2-4} 
   \multicolumn{1}{|c|}{} &
     \multicolumn{1}{c|}{Acc$\uparrow$} &
     \multicolumn{1}{c|}{ECE$\downarrow$} &
     \multicolumn{1}{c|}{AECE$\downarrow$}  \\ \hline
     \multicolumn{1}{|l|}{CE }                             & 73.96              & 9.10                  & 9.24    \\
     \multicolumn{1}{|l|}{Mixup~\cite{thulasidasan2019mixup} }                                     & \textbf{75.84}            & 7.07                 &7.09              \\
     \multicolumn{1}{|l|}{CRL~\cite{moon2020confidence} }                                     & 73.83           & 8.47              & 8.47             \\
     \multicolumn{1}{|l|}{MbLS~\cite{liu2022devil}     }                               &  75.39            & \underline{4.07} & \underline{4.14}  \\ 
     \multicolumn{1}{|l|}{RegMixup~\cite{pinto2022using} }                              & \underline{75.64}            & 5.34                 & 5.42            \\ \hline
     \multicolumn{1}{|l|}{RankMixup~(MRL) }                                                       & 75.27         & 5.70             & 5.71   \\ 
     \multicolumn{1}{|l|}{RankMixup~(M-NDCG) }                                                       & 74.86         & \textbf{3.93}     & \textbf{3.92}  \\ \hline
   \end{tabular}
}

\end{table}

\subsection{Results}\label{sec:results}

\paragraph{Comparison with the state of the art.}
We compare in Table~\ref{tab:result} our method with state-of-the-art calibration approaches. From the tables, we observe four things: (1) Our models outperform other mixup-based methods~\cite{pinto2022using,thulasidasan2019mixup} using label mixtures as supervisions, demonstrating the effectiveness of our approach to leveraging an ordinal ranking relationship for confidences. (2) Our model using M-NDCG achieves the best performance in terms of both ECE and AECE on the challenging Tiny-ImageNet~\cite{le2015tiny} dataset, while providing competitive results on CIFAR10/100. This suggests that the ordinal ranking relationships among confidences is crucial for network calibration, and M-NDCG effectively encourages our model to provide confidences maintaining the relationships. We can also see that our method generalizes well to the large-scale dataset. (3) Our method outperforms another ranking-based method~(CRL)\cite{moon2020confidence} significantly, suggesting that more complex ranking relationship based on augmented samples enables estimating better confidences. (4) Ours with the post-hoc method~\cite{guo2017calibration} gives the best results in most settings, which indicates that they are complementary to each other, and exploiting both further boosts the calibration accuracy. 

We also provide in Tables~\ref{tab:imagenet} and~\ref{tab:wideresnet} quantitative results on ImaegeNet~\cite{deng2009imagenet} and for the WideResNet backbone~\cite{zagoruyko2016wide}, respectively. They show similar trends to those in Table~\ref{tab:result}, validating the effectiveness of our approach across various network architectures and large-scale datasets. Specifically, our models provide competitive results with the state of the art~\cite{liu2022devil}, while outperforming mixup-based methods~\cite{thulasidasan2019mixup,pinto2022using} and the ranking-based one~\cite{moon2020confidence}.

\begin{table}[]

   \centering
   \vspace{-2mm}

   \captionsetup{font={small}}
      \caption{Quantitative comparison in terms of ECE~(\%) and AECE~(\%) with 15 bins on the validation splits of CIFAR10~\cite{krizhevsky2009learning} and Tiny-ImageNet~\cite{le2015tiny}. Numbers in bold indicate the best performance and underscored ones are the second best.}
      \vspace{-2mm}

   \label{tab:wideresnet}
   \setlength{\tabcolsep}{1.3pt}
   \renewcommand{\arraystretch}{1.0}
   {\scriptsize
           \begin{tabular}{l|ccc|ccc|}
            \cline{2-7}
            & \multicolumn{3}{c|}{CIFAR10~\cite{krizhevsky2009learning}} & \multicolumn{3}{c|}{Tiny-ImageNet~\cite{le2015tiny}}  \\ \hline
            \multicolumn{1}{|c|}{\multirow{2}{*}{Methods}} &
              \multicolumn{6}{c|}{Wide-ResNet-26-10~\cite{zagoruyko2016wide}}  \\ \cline{2-7} 
            \multicolumn{1}{|c|}{} &
              \multicolumn{1}{c|}{Acc$\uparrow$} &
              \multicolumn{1}{c|}{ECE$\downarrow$} &
              \multicolumn{1}{c|}{AECE$\downarrow$} &
              \multicolumn{1}{c|}{Acc$\uparrow$} &
              \multicolumn{1}{c|}{ECE$\downarrow$} &
              \multicolumn{1}{c|}{AECE$\downarrow$}\\ \hline 
   \multicolumn{1}{|l|}{CE }                             & \underline{95.80}              & 2.70                  & 2.66    & 65.18              & 6.08                  & 6.06  \\
   \multicolumn{1}{|l|}{Mixup~\cite{thulasidasan2019mixup} }                                     & \textbf{96.53}            & 3.14                 &3.08   & \textbf{66.36}            & 3.77               &3.75             \\
   \multicolumn{1}{|l|}{MbLS~\cite{liu2022devil}}                                    &  95.70           & \textbf{1.45} & 2.78 &  65.30            & \textbf{2.57} & \textbf{2.32}  \\ 
   \multicolumn{1}{|l|}{RegMixup~\cite{pinto2022using} }                              & 95.44            & 4.18                 &3.99      &63.40            & 3.87                 & 3.93        \\ \hline
   \multicolumn{1}{|l|}{RankMixup~(MRL)}                                                        & 95.57      & 1.70             & \textbf{1.38}   & 65.34      &   \underline{3.30}         & \underline{3.28} \\ 
   \multicolumn{1}{|l|}{RankMixup~(M-NDCG)  }                                                      & 95.73         & \underline{1.62}             & \underline{1.53} & \underline{65.56}          & 3.83             & 3.94  \\ \hline
            
           \end{tabular}%
   }

   \end{table}
   
   \begin{table}[]
      \vspace{-2mm}
   
      \centering
      \captionsetup{font={small}}
      \caption{Quantitative comparison with the state of the art in terms of AUROC~(\%). Numbers in bold indicate the best performance and underscored ones indicate the second best.}
      \vspace{-2mm}
   
      \label{tab:ood}
      \setlength{\tabcolsep}{1.2pt}
      \renewcommand{\arraystretch}{1.0}
      {\scriptsize
      \resizebox{1.0\columnwidth}{!}{%
      \begin{tabular}{|l|ccc|ccc|}
      \hline
       \multicolumn{1}{|c|}{ID} & 
       \multicolumn{3}{c|}{C10~\cite{krizhevsky2009learning}} &
       \multicolumn{3}{c|}{C100~\cite{krizhevsky2009learning}}    
       \\ \hline
       \multicolumn{1}{|c|}{\multirow{2}{*}{OOD}} &
        \multicolumn{1}{c|}{C100~\cite{krizhevsky2009learning}} &
        \multicolumn{1}{c|}{Tiny~\cite{le2015tiny}} &
        \multicolumn{1}{c|}{SVHN~\cite{netzer2011reading}} &
        \multicolumn{1}{c|}{C10~\cite{krizhevsky2009learning}} &
        \multicolumn{1}{c|}{Tiny~\cite{le2015tiny}} &
        \multicolumn{1}{c|}{SVHN~\cite{netzer2011reading}} \\\cline{2-7}
        \multicolumn{1}{|c|}{}&
        \multicolumn{3}{c|}{AUROC$\uparrow$} &
        \multicolumn{3}{c|}{AUROC$\uparrow$}\\
        \hline
      TS~\cite{guo2017calibration}                              & 86.73             & 88.06                 & 92.14      & 76.38              & 80.12             & 82.41 \\
      LS~\cite{muller2019does}                             & 75.23       & 76.29 & 74.23 & 72.92             & 77.54 & 72.33            \\
      FL~\cite{mukhoti2020calibrating}                                      & 85.63            & 86.84 & \underline{92.66} & 78.37             & \underline{81.23} & \bf{86.37}             \\
      Mixup~\cite{thulasidasan2019mixup}                                      & 82.07              & 84.09                  & 79.72      & \underline{78.46}             & \bf{81.55}             & 76.21           \\
      CPC~\cite{cheng2022calibrating}                             & 85.28 & 85.15 & \bf{92.79} & 74.68        & 76.66 & 71.72            \\
      MbLS~\cite{liu2022devil}                                    &  86.20             & 87.55 &89.98 & 73.88             & 79.19 & 73.19 \\ 
      RegMixup~\cite{pinto2022using}                               & \textbf{87.82}             & 87.54                  & 83.15      & 76.00             &80.72             & 75.05        \\ \hline
      RankMixup~(M-NDCG)                                                        & \textbf{87.82}         & \bf{88.94}             & 92.62 & \bf{78.64}              & 80.67        & \underline{85.42} \\ \hline
      \end{tabular}%
      }
      }
      \vspace{-3mm}
   
      \end{table}
\vspace{-2mm}
\paragraph{OOD detection.}
We compare in Table~\ref{tab:ood} the OOD detection accuracy using the ResNet-50~\cite{he2016deep} architecture. We use the entropy of the softmax output as an uncertainty measure for binary classification and report AUROC scores for comparison. From the table, we have two findings. Our model trained with M-NDCG outperforms other mixup-based methods and achieves the highest AUROC scores in most settings. This indicates that ordinal ranking relationship of confidences is also crucial for OOD detection, suggesting that models trained with M-NDCG are not only better calibrated but also perform better than other calibration methods under the distributional shifts.

\begin{table}[]
   \vspace{-2mm}
   \centering

   \captionsetup{font={small}}
   \captionof{table}{Quantitative comparison by varying the margin~$m$. We report the ECE~(\%) for the models trained with MRL. Numbers in parentheses represent the results obtained using a TS post-hoc technique~\cite{guo2017calibration}.}
   \label{tab:margin}
   \setlength{\tabcolsep}{1.4pt}
\renewcommand{\arraystretch}{1.1}
{\scriptsize
     \centering
     \begin{tabular}{c|c |c|}
      \cline{2-3} 
               & CIFAR10~\cite{krizhevsky2009learning} & Tiny-ImageNet~\cite{le2015tiny} \\ \cline{2-3}
              &       ECE$\downarrow$    &     ECE$\downarrow$           \\ \hline
              \multicolumn{1}{|l|}{$m=1.0$}      & 1.54(1.09)& 1.49(1.49)  \\
              \multicolumn{1}{|l|}{$m=2.0$} & 1.01(0.84)& 1.70(1.70) \\ 
              \multicolumn{1}{|l|}{$m=3.0$ } & 2.12(0.93)& 1.54(1.54)\\
              \multicolumn{1}{|l|}{$m=4.0$} & 3.12(1.67)&2.37(1.77)\\
              \multicolumn{1}{|l|}{ $m=5.0$} & 13.13(0.48)&2.42(1.48) \\ \hline
     \end{tabular}
}
   \end{table}

\subsection{Discussion}\label{sec:discussion}
In this section, we analyze the effects of the hyperparameters in our framework, and show generalization ability by incorporating other mixup techniques into our framework. For the analyses, we mainly adopt the ResNet-50~\cite{he2016deep} models trained with M-NDCG, and report ECE and AECE on the validation sets of CIFAR10~\cite{krizhevsky2009learning} and Tiny-ImageNet~\cite{le2015tiny}, unless otherwise specified. We use 15 bins to measure the ECE and AECE.
\vspace{-2mm}

\begin{figure}[t]
   \vspace{-2mm}

   \captionsetup{font={small}}
      \centering
         \includegraphics[width=0.8\linewidth]{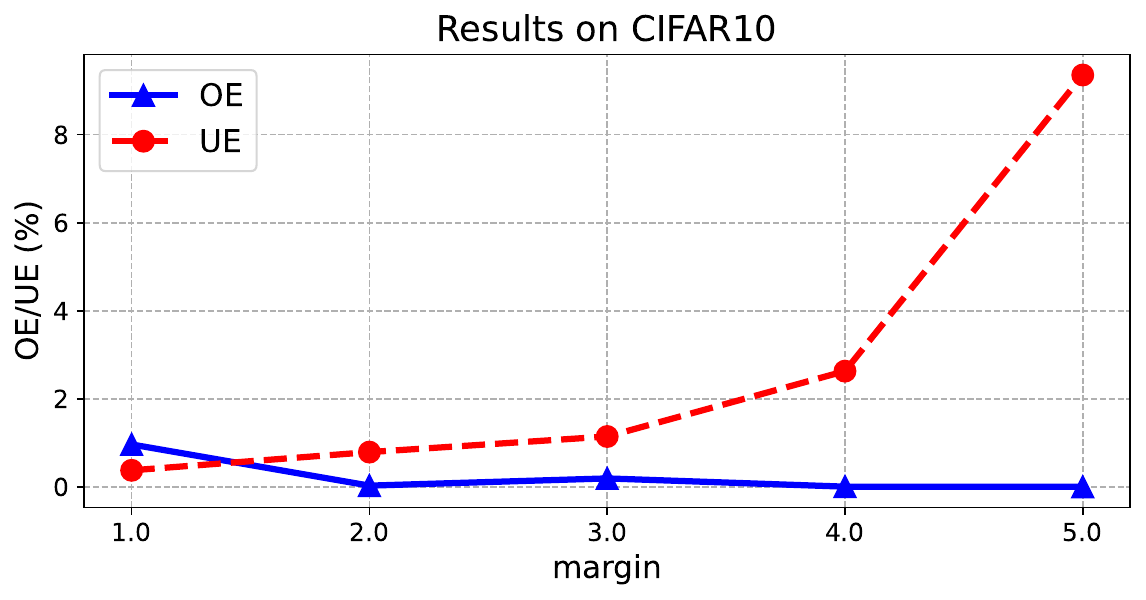}
         \vspace{-3mm}

         \caption{Quantitative results by varying the margin~$m$. We plot OE~(\%) and UE~(\%) on the validation split of CIFAR10~\cite{krizhevsky2009learning}. Best viewed in color.} 
   \vspace{-2mm}

   \label{fig:oeue}
   \end{figure}
\vspace{-2mm}
   
\paragraph{Margin.}
We compare in Table~\ref{tab:margin} the calibration results of the models trained with MRL and various margins. We can see that small margins provide better performances for both datasets. Furthermore, we find that increasing the margin degrades the classification accuracy. A reason is that a large margin could make a significant difference in confidence levels between the raw and augmented samples. That is, with a larger margin, the augmented sample is more likely to have a lower confidence than its actual value.

 To demonstrate this, we compute overconfidence errors~(OE)~\cite{thulasidasan2019mixup} that measure the expected difference between confidence values of overconfident predictions and corresponding accuracies. Given $H$ bins of equal size, OE is defined as follows~\cite{thulasidasan2019mixup}:
\begin{equation}\scalemath{0.9}{
   OE =\sum_{h=1}^{H}\frac{\left|B_{h}\right|}{N}\left[\mathrm{conf}(B_{h})\times \max(\mathrm{conf}(B_{h})-\mathrm{acc}(B_{h}),0)\right],}	
\end{equation}
where $N$ is the total number of samples, and $B_{h}$ is the set of samples, whose confidences fall into the $h$-th bin. We denote by $\mathrm{acc}(B_{h})$ and $\mathrm{conf}(B_{h})$ the accuracy and averaged confidence of the $h$-th bin, respectively. Similar to OE, we define underconfidence errors~(UE) computing the difference using underconfident predictions as follows: 
\begin{equation}\scalemath{0.9}{
   UE =\sum_{h=1}^{H}\frac{\left|B_{h}\right|}{N}\left[\mathrm{conf}(B_{h})\times \max(\mathrm{acc}(B_{h})-\mathrm{conf}(B_{h}),0)\right].}	
\end{equation}
We plot in Fig.~\ref{fig:oeue} UE and OE for ResNet-50 with MRL on CIFAR10~\cite{krizhevsky2009learning}, while varying margins. We can see that UE increases drastically, according to the margins, while OE decreases gradually. This suggests that exploiting a large margin causes underconfidence problems, where the model becomes less confident in its predictions. An underconfident model produces large calibration errors~\cite{thulasidasan2019mixup,zhang2018mixup}, leading to the degraded performance.

   \begin{table}[]
      \centering

   \vspace{-2mm}
   \captionsetup{font={small}}
   \captionof{table}{Quantitative results of our model with various numbers of the augmented samples. We report the ECE~(\%) for comparison. Numbers in parentheses represent the results obtained using a TS post-hoc technique~\cite{guo2017calibration}.}
   \label{tab:Qnum}
   \setlength{\tabcolsep}{1.4pt}
   \renewcommand{\arraystretch}{1.0}
   {\scriptsize
     \begin{tabular}{c|c| c|}
     \cline{2-3}
              & CIFAR10~\cite{krizhevsky2009learning} & Tiny-ImageNet~\cite{le2015tiny} \\ \cline{2-3}
              &        ECE$\downarrow$    &     ECE$\downarrow$          \\ \hline
              \multicolumn{1}{|l|}{$Q=2$}      & 3.01(0.82)& 3.25(1.44) \\
              \multicolumn{1}{|l|}{$Q=3$} & 2.78(0.50)& 2.18(1.12)  \\ 
              \multicolumn{1}{|l|}{$Q=4$}  & 2.59(0.57)& 1.49(1.49)\\
              \multicolumn{1}{|l|}{$Q=5$} & 2.37(0.64) & 1.52(1.52)\\
              \multicolumn{1}{|l|}{$Q=6$} & 2.24(0.59)& 1.30(1.30)\\ \hline
      
     \end{tabular}
}
\vspace{-2mm}

   \end{table}

\vspace{-2mm}

\paragraph{Number of augmented samples.}
We compare in Table~\ref{tab:Qnum} the calibration performance in terms of ECE with varying number of augmented samples. We set $\alpha$ to 2.0 and 1.0 for CIFAR10 and Tiny-ImageNet, respectively. We can observe that ECE improves as more augmented samples are used, suggesting that learning with diverse ranking relationships between a number of augmented samples encourages the model to estimate confidence levels better. This further implies that the mixup coefficient~$\lambda$ is also an important factor, since it enables modeling the diverse ranking relationships. This leads us to analyze the mixup coefficient~$\lambda$ in detail in the following.

\begin{figure}[t]
   \vspace{-2mm}
   \captionsetup{font={small}}
      \centering
         \includegraphics[width=0.8\linewidth]{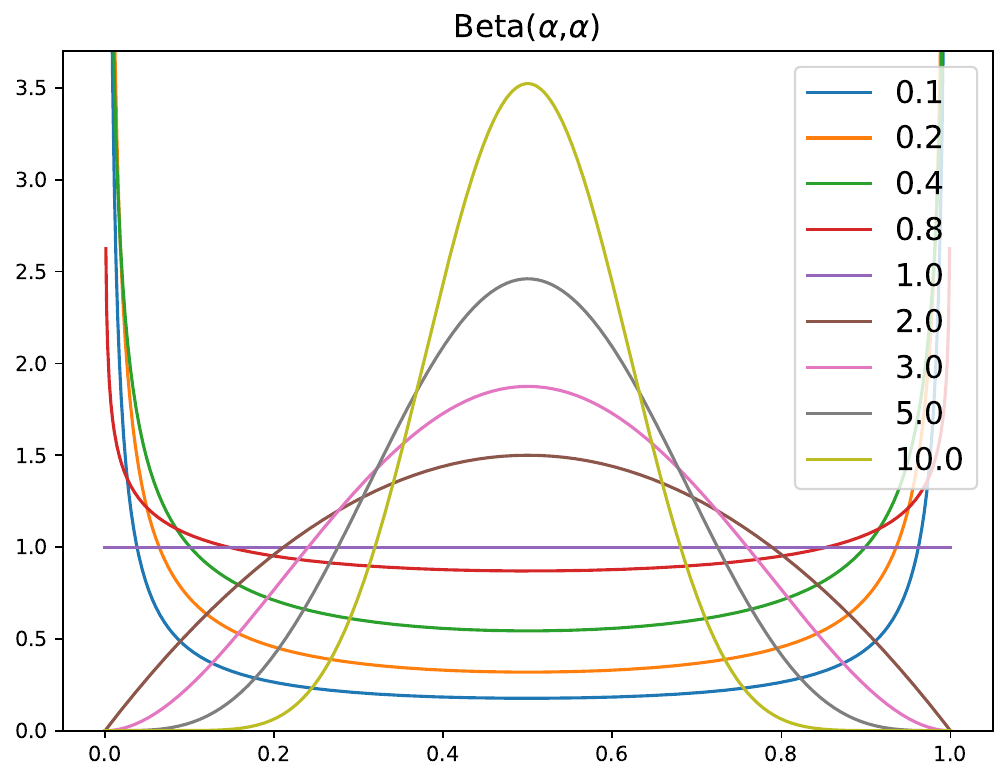}
\vspace{-0.3cm}
 \caption{Probability density function~(PDF) of \textit{Beta} distribution with various $\alpha$s.} 
  \label{fig:beta}
  \vspace{-4mm}
  \end{figure}

  \vspace{-2mm}

\begin{figure}[t]
   \vspace{-2mm}
   \begin{center}
   \begin{subfigure}{0.8\columnwidth}
   \includegraphics[width=\textwidth]{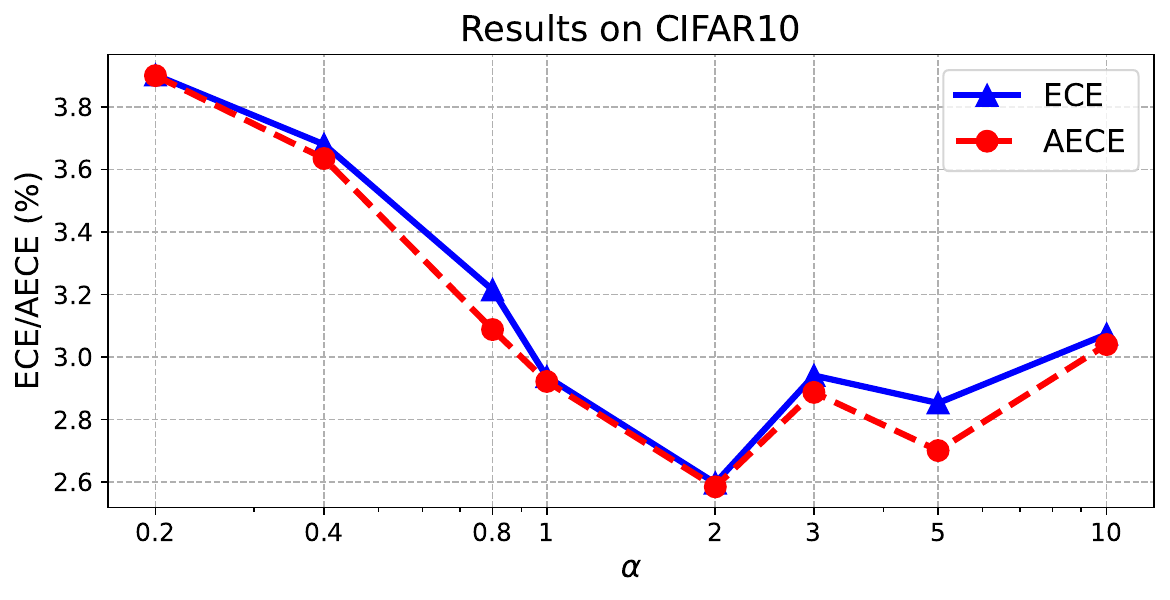}
   \end{subfigure}
   \begin{subfigure}{0.8\columnwidth}
   \includegraphics[width=\textwidth]{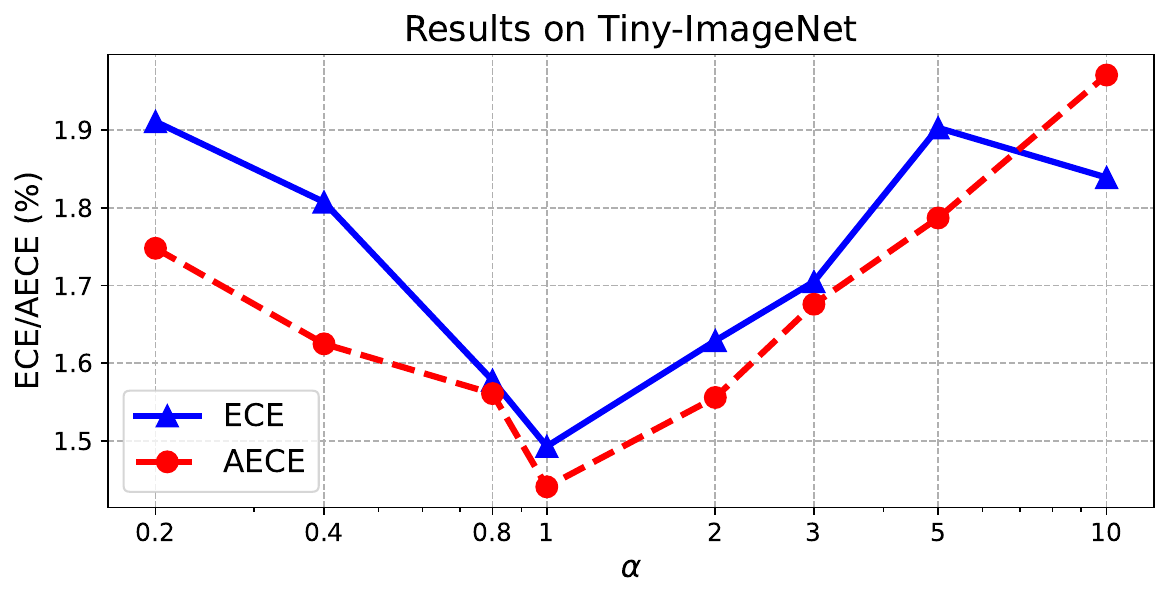}
   \end{subfigure}
\end{center}
\vspace{-0.7cm}
\captionsetup{font={small}}
\caption{Quantitative results by varying the parameter of the Beta distribution~($\alpha$). We plot ECE~(\%) and AECE~(\%) on the validation split of CIFAR10~\cite{krizhevsky2009learning}~(\textit{top}) and Tiny-ImageNet~\cite{le2015tiny}~(\textit{bottom}). Best viewed in color.}
\label{fig:alpha}
\vspace{-4mm}

\end{figure}

  \paragraph{Mixup coefficient.}
  Following~\cite{zhang2018mixup}, we sample the mixup coefficient~$\lambda$ from a \textit{Beta} distribution, making it possible to generate various augmented samples. Note that the degree of the diversity is determined by how we set the shape parameter for the \textit{Beta} distribution. For an in-depth analysis, we also show in Fig.~\ref{fig:beta} the probability density functions~(PDFs) of different $Beta(\alpha,\alpha)$ distributions, while varying the shape parameter~$\alpha$. We evaluate the calibration errors of the models trained with M-NDCG along with different shape parameters~$\alpha$ of the \textit{Beta} distribution, and show the results using 3 augmented images~($Q=4$) in Fig.~\ref{fig:alpha}. From the figures, we observe two things: (1) Our models achieve the best performances with $\alpha=2$ and $\alpha=1$ on CIFAR10 and Tiny-ImageNet, respectively, where the corresponding PDFs are relatively flat, as shown in Fig.~\ref{fig:beta}. This suggests that diverse mixup coefficients~$\lambda$ drawn from such distributions can enhance the calibration performances, as it allows us to augment the samples in a more diverse way. (2) Our model performs well with large~$\alpha$, which is in contrast to the vanilla mixup~\cite{zhang2018mixup,thulasidasan2019mixup}, where the value of~$\alpha$ within the range of $[0.1, 0.4]$ yields better classification and calibration performances. When $\alpha$ is too large, it suffers from an underconfidence problem~\cite{zhang2018mixup,thulasidasan2019mixup} and is also affected by manifold intrusion~\cite{guo2019mixup}, leading to significant performance drops in both the classification and calibration. The manifold intrusion happens when the label mixture of an augmented sample differs from the label of the real sample~\cite{guo2019mixup}. However, in our framework, we do not explicitly use the label mixtures as supervision, suggesting that our framework is less affected by those problems. Therefore, our framework is more robust to the variation of the parameter~$\alpha$ compared to the vanilla mixup. To further investigate the importance of diverse samples, we also perform experiments with various combinations of $\alpha$ and $Q$ and provide the results in the supplementary material.

  \vspace{-2mm}

\begin{table}[t]
   \vspace{-2mm}
   \captionsetup{font={small}}
   \caption{Quantitative comparison of calibration performances using other mixup techniques on the validation splits of CIFAR10~\cite{krizhevsky2009learning} and Tiny-ImageNet~\cite{le2015tiny}.  }
   \vspace{-2mm}

   \centering
   \label{tab:variants}
   \setlength{\tabcolsep}{1.0pt}
   \renewcommand{\arraystretch}{1.1}
   \resizebox{1\columnwidth}{!}{%
   {\scriptsize
   \begin{tabular}{l|ccc |c|c |c|c|}
   \cline{2-8}
   &\multirow{2}{*}{\shortstack{Mixup~\cite{zhang2018mixup}}}& \multirow{2}{*}{\shortstack{Manifold~\cite{verma2019manifold}}} & \multirow{2}{*}{CutMix~\cite{yun2019cutmix}} & \multicolumn{2}{c|}{CIFAR10~\cite{krizhevsky2009learning}} &   \multicolumn{2}{c|}{Tiny-ImageNet~\cite{le2015tiny}} \\ \cline{5-8}
                        & &                                                                         &                                              & ECE$\downarrow$              & AECE$\downarrow$      & \hspace{0.1cm}ECE$\downarrow$   & \hspace{0.1cm}AECE$\downarrow$ \\\hline
                        \multicolumn{1}{|l|}{\multirow{2}{*}{RegMixup~\cite{pinto2022using}} } & &$\checkmark$                                 &                                              & 3.63             & 3.57      & 4.76  & 4.88 \\\multicolumn{1}{|l|}{}
                                                    &&                                              & $\checkmark$                                 & 3.56 & 3.54        & 2.97  & 2.93 \\ \hline
                                                    \multicolumn{1}{|l|}{\multirow{3}{*}{RankMixup} }                        & & $\checkmark$                                 &                                              & 2.92             & 2.98      & 1.20  & 1.22  \\\multicolumn{1}{|l|}{}
                                                    &&                                              & $\checkmark$                                 & 3.26             & 3.21      & 1.80   & 1.66 \\ \multicolumn{1}{|l|}{}
                     &                 $\checkmark$               &                                              &                                  & 2.59             & 2.58      & 1.49   & 1.44 \\\hline
   \end{tabular}%
   }
   }
   \vspace{-2mm}

\end{table}

\begin{table}[]

   \captionsetup{font={small}}
   
   \caption{Quantitative comparison of classification and calibration performances on the test splits of CIFAR10-LT~\cite{cao2019learning} and CIFAR100-LT~\cite{cao2019learning}. We report the Top-1 accuracy~(\%) and ECE~(\%) with 15 bins.}
   \vspace{-2mm}
   \centering
   \label{tab:lt}
   \setlength{\tabcolsep}{1.3pt}
   \renewcommand{\arraystretch}{1.1}
   {\scriptsize
   \begin{tabular}{l |cc |cc|}
   \cline{2-5}
   &
   \multicolumn{2}{c|}{CIFAR10-LT~\cite{cao2019learning}} &   \multicolumn{2}{c|}{CIFAR100-LT~\cite{cao2019learning}} \\ \hline
   \multicolumn{1}{|c|}{\multirow{2}{*}{Methods}}
   & $\rho=10$              & $\rho=100$      & $\rho=10$   & $\rho=100$ \\\multicolumn{1}{|l|}{}
   &\hspace{0.1cm}Acc$\uparrow$/ECE$\downarrow$ & \hspace{0.1cm}Acc$\uparrow$/ECE$\downarrow$ & \hspace{0.1cm}Acc$\uparrow$/ECE$\downarrow$ & \hspace{0.1cm}Acc$\uparrow$/ECE$\downarrow$ \\\hline
   \multicolumn{1}{|l|}{ CE }                             & 86.39/6.60              & 70.36/20.53                  & 55.70/22.85    & 38.32/38.23               \\
   \multicolumn{1}{|l|}{Mixup~\cite{zhang2018mixup} }                                     & 87.10/6.55           & 73.06/19.20                 &58.02/19.69   & 39.54/32.72                  \\
   \multicolumn{1}{|l|}{Remix~\cite{chou2020remix}  }                                    & 88.15/6.81           & 75.36/15.38                 &59.36/20.17   & 41.94/33.56                  \\
   \multicolumn{1}{|l|}{UniMix~\cite{xu2021towards}   }                           & 89.66/6.00           & 82.75/12.87                 & 61.25/19.38     & 45.45/27.12                 \\\hline
   \multicolumn{1}{|l|}{RankMixup~(MRL)  }                             & 88.83/2.16           & 74.16/\hspace{0.1cm}2.03                 & 62.84/13.55     & 42.56/\hspace{0.1cm}5.28  \\
   \multicolumn{1}{|l|}{RankMixup~(M-NDCG) }                              & 89.80/5.94           & 75.41/14.10                 & 63.83/\hspace{0.1cm}9.99     & 43.00/18.74  \\\hline
   \end{tabular}%
   }
   \vspace{-2.5mm}

\end{table}

\paragraph{Other mixup techniques.}
RankMixup employs a vanilla mixup technique that interpolates the input images and the corresponding labels for data augmentation, but other variants can be readily adopted in our framework. We have also tried using Manifold-Mixup~\cite{verma2019manifold} and CutMix~\cite{yun2019cutmix} for the experiments. We show in Table~\ref{tab:variants} the results of our models using these techniques together with M-NDCG, while setting the shape parameter $\alpha$ to 1 for the both techniques. For comparison, we report the result of our model with the vanilla mixup, which exploits 3 augmented samples, and additionally show the performances obtained using RegMixup~\cite{pinto2022using} equipped with these variants. For both datasets, our method achieves better calibration errors than RegMixup, indicating that our method has better generalization capability. While our model with vanilla mixup provides the best performance on CIFAR10, employing Manifold-Mixup offers better results on Tiny-ImageNet in terms of calibration errors. Increasing the entropy of supervision is crucial for better-calibrated models~\cite{muller2019does,thulasidasan2019mixup,liu2022devil,mukhoti2020calibrating}. RankMixup exploits the augmented samples as supervision for network calibration, indicating that samples with higher entropies can yield better performances. Manifold-Mixup~\cite{verma2019manifold} produces flattened representations that have higher entropies than the mixup-augmented samples, which may explain the reason why our model with Manifold-Mixup performs better on Tiny-ImageNet.  

\vspace{-3mm}

\paragraph{Long-tailed datasets.}
We evaluate our method on long-tailed~(LT) datasets, where the miscalibration issues of DNNs are more pronounced~\cite{zhong2021improving,xu2021towards}. Following the experimental settings in~\cite{zhong2021improving,xu2021towards}, we adopt the ResNet32~\cite{he2016deep} architecture as our backbone, and exploit CIFAR10/100-LT datasets~\cite{cao2019learning} with two imbalanced factors~($\rho$). We compare in Table~\ref{tab:lt} our method against the vanilla mixup~\cite{zhang2018mixup} and two adaptive mixup techniques tailored for LT scenarios~\cite{chou2020remix,xu2021towards}. We can see that our models significantly outperform other mixup methods in terms of ECE, while providing competitive accuracies, although they are not specially designed for LT scenarios. This suggests that leveraging the ordinal ranking relationship between confidences is also effective for addressing miscalibration problems in LT scenarios, instead of using label mixtures for other mixup-based augmentations~\cite{zhong2021improving,xu2021towards}. Note that the number of samples for head and tail classes is highly skewed, suggesting that augmented samples are not diverse in LT scenarios.  As this problem is more severe for M-NDCG, it yields worse results, compared to MRL, in the LT datasets.

\section{Conclusion}
We have presented a novel mixup-based framework for network calibration, dubbed RankMixup, alleviating the problem that interpolated labels do not necessarily model the actual distribution of label mixtures. To this end, we have proposed a new supervisory signal based on the ordinal ranking relationship that a higher confidence is more favorable for a raw sample than the one augmented by mixup. We have proposed MRL, which encourages a network to yield confidences that preserve the ranking relationships between the raw and mixup-augmented samples. We have also extended this idea to multiple augmented samples with different mixup coefficients via M-NDCG, enabling a network to predict outputs with higher confidences for augmented samples that are closer to a raw one. We have demonstrated the effectiveness of RankMixup through the extensive experiments and analysis, achieving the best results among mixup-based network calibration approaches.

{\small
\paragraph{Acknowledgments.}This work was supported in part by the NRF and IITP grants funded by the Korea government~(MSIT) (No.2023R1A2C2004306, No.2022-0-00124, Development of Artificial Intelligence Technology for Self-Improving Competency-Aware Learning Capabilities, No.2021-0-02068, Artificial Intelligence Innovation Hub), and the Yonsei Signature Research Cluster Program of 2023~(2023-22-0008).
}

{\small
\bibliographystyle{ieee_fullname}
\bibliography{egbib}
}

\clearpage


\section*{Supplementary material}

# Supplementary Material
# RankMixup: Ranking-based Mixup Training for Network Calibration

Jongyoun Noh    Hyekang Park    Junghyup Lee    Bumsub Ham*

School of Electrical and Electronic Engineering, Yonsei University

## 1. Implementation details

**Methods for comparison.** We provide details of the hyperparameters used for reproducing the results of other methods as follows: (a) ECP [9]: We set the weight of entropy penalty to 0.1 following [9] (b) LS [8]: Following [5, 7], we report the results obtained with $\alpha = 0.05$. (c) FL [7]: We train the models with a fixed regularization parameter ($\gamma$) of 3. (d) Mixup [11]: We use a shape parameter $\alpha$ of 0.2, the best performing one in [11]. (e) FLSD [7]: Following the schedule in [7], we use the parameter $\gamma$ of 5 for the samples whose output probability for the ground-truth class is within $[0, 0.2)$, otherwise we use the parameter $\gamma$ of 3. (f) CRL [6]: We set the balancing parameter of the ranking loss as 1.0. (g) CPC [1]: We set the weights of binary discrimination and binary exclusion losses as 0.1 and 1.0, respectively. (h) MbLS [5]: We train the models with margins of 6 and 10 for the CIFAR10/100 datasets and Tiny-ImageNet dataset, respectively. (i) RegMixup [10]: We obtain the results with a shape parameter $\alpha$ of 10.0 as suggested in [10].

## 2. More results

**Q and $\alpha$.** We perform experiments with various combinations of $\alpha$ and $Q$ to further investigate the importance of diverse samples. We adopt the ResNet-50 [2] models trained with M-NDCG for the analyses, and show the calibration performances in terms of ECE and AECE on the validation sets of CIFAR10 [3] and Tiny-ImageNet [4] in Fig. 1 and 2, respectively. From the figures, we can observe three things: (1) Our models using $\alpha$ within the range of $[1, 3]$ achieve the best performances in most cases. As shown in Fig. 2 of the main paper, such distributions enable the model to sample more diverse and large mixup coefficients $\lambda$, suggesting that generating diverse samples with relatively strong interpolations is crucial for our framework. (2) Our models also perform better with larger $\alpha$s than smaller ones, in contrast to vanilla mixup [11, 12]. However, in some cases, the performances worsen with very large $\alpha$. (3) For most $\alpha$ values, both ECE and AECE improve as more augmented samples are used. This indicates that incorporating more diverse ranking relationships based on augmented samples is favorable in our framework. However, using more samples with very small or large $\alpha$ leads to degraded performance. Overall, these observations highlight the importance of using diverse samples and considering a range of $\alpha$ values for effective calibration in our framework.

---

*Corresponding author.

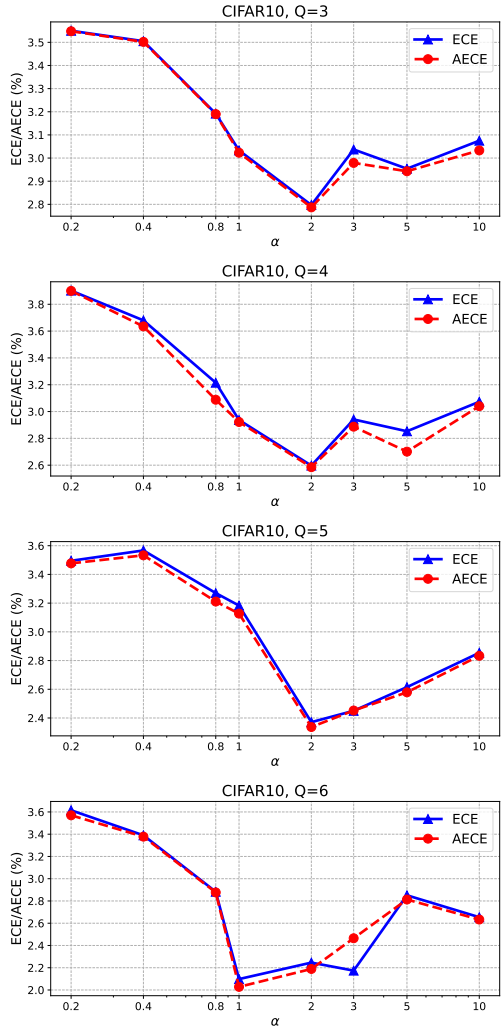

Figure 1: Quantitative results with various combinations of the parameter of *Beta* distribution ($\alpha$) and the number of aumgented samples ($Q$). We plot the variation of both ECE (%) and AECE (%) on the validation split of CIFAR10 [3]. Best viewed in color.

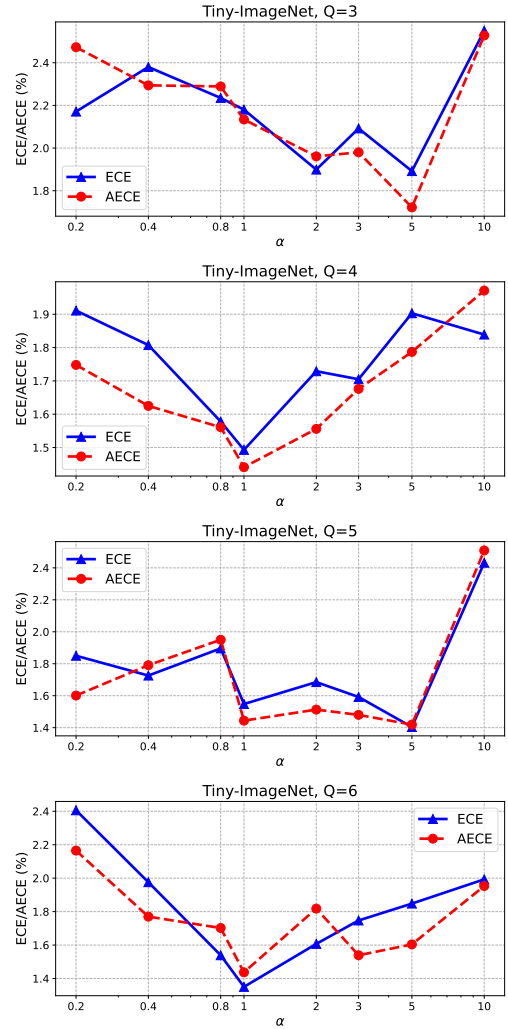

Figure 2: Quantitative results with various combinations of the parameter of *Beta* distribution ($\alpha$) and the number of aumgented samples ($Q$). We plot the variation of both ECE (%) and AECE (%) on the validation split of Tiny-ImageNet [4]. Best viewed in color.



\end{document}